\documentclass[twoside,11pt]{article}
\usepackage{jair, theapa, rawfonts}

\usepackage{caption} 
\usepackage{times}
\usepackage{soul}
\usepackage{url}
\usepackage[utf8]{inputenc}
\usepackage{caption}
\usepackage{float}
\usepackage{subfig}
\usepackage{graphicx}
\usepackage{lscape}

\usepackage{amsmath}
\usepackage{amssymb}
\usepackage{booktabs}
\usepackage{algorithm}
\usepackage{algorithmic}
\usepackage{tikz}
\usepackage[edges]{forest}
\usetikzlibrary{shadows}

\newcommand{\bfx}{\mathbf{x}}

\begin{document}

\title{Calibration in Deep Learning: A Survey of the State-of-the-Art}

\author{\name Cheng Wang \email cwngam@amazon.com  \\
       \addr Amazon }


\maketitle

\begin{abstract}
Calibrating deep neural models plays an important role in building reliable, robust AI systems in safety-critical applications. Recent work has shown that modern neural networks that possess high predictive capability are poorly calibrated and produce unreliable model predictions. Though deep learning models achieve remarkable performance on various benchmarks, the study of model calibration and reliability is relatively under-explored. Ideal deep models should have not only high predictive performance but also be well calibrated. There have been some recent advances in calibrating deep models. In this survey, we review the state-of-the-art calibration methods and their principles for performing model calibration. First, we start with the definition of model calibration and explain the root causes of model miscalibration. Then we introduce the key metrics that can measure this aspect. It is followed by a summary of calibration methods that we roughly classify into four categories: post-hoc calibration, regularization methods, uncertainty estimation, and composition methods. We also cover recent advancements in calibrating large models, particularly large language models (LLMs). Finally, we discuss some open issues, challenges, and potential directions.
\end{abstract}

\section{Introduction}

Deep Neural Networks (DNNs) have been showing promising predictive power in many domains such as computer vision~\cite{krizhevsky2012imagenet}, speech recognition~\cite{graves2013speech} and natural language processing~\cite{vaswani2017attention}. Nowadays, deep neural network models are frequently being deployed into real-world systems. However, recent work~\cite{guo2017calibration} pointed out that those highly accurate, negative-log-likelihood (NLL) trained deep neural networks are poorly calibrated~\cite{niculescu2005predicting}, i.e., the model predicted class probabilities do not faithfully estimate the true correctness likelihood and lead to overconfident and underconfident predictions. 

Although in production environments, the primary goal of deep learning models is to make accurate predictions that can inform decision-making processes, in many real-world applications, decisions are not based solely on whether a prediction is correct or incorrect, but also on the confidence associated with that prediction. 
Deploying uncalibrated models into real-world systems is at high risk, particularly for safety-critical applications such as medical diagnosis~\cite{caruana2015intelligible}, autonomous driving~\cite{bojarski2016end} and finance decision-making~\cite{buchel2022deep}. One concrete example is the decision making based on model predictions in a medical diagnostic system. The model might predict that a patient has a 70\% chance of having a certain disease based on their symptoms and medical history. A well-calibrated model ensures that if it predicts a 70\% chance of an event happening, that event indeed occurs approximately 70\% of the time in reality. Without calibration, the reliability of model predictions is compromised, leading to inaccurate assessments of uncertainty and potentially flawed decision-making.

\begin{figure}[!htb]
    \centering
    \subfloat[Reliability diagram (uncalibrated)]{\includegraphics[width=0.325\textwidth]{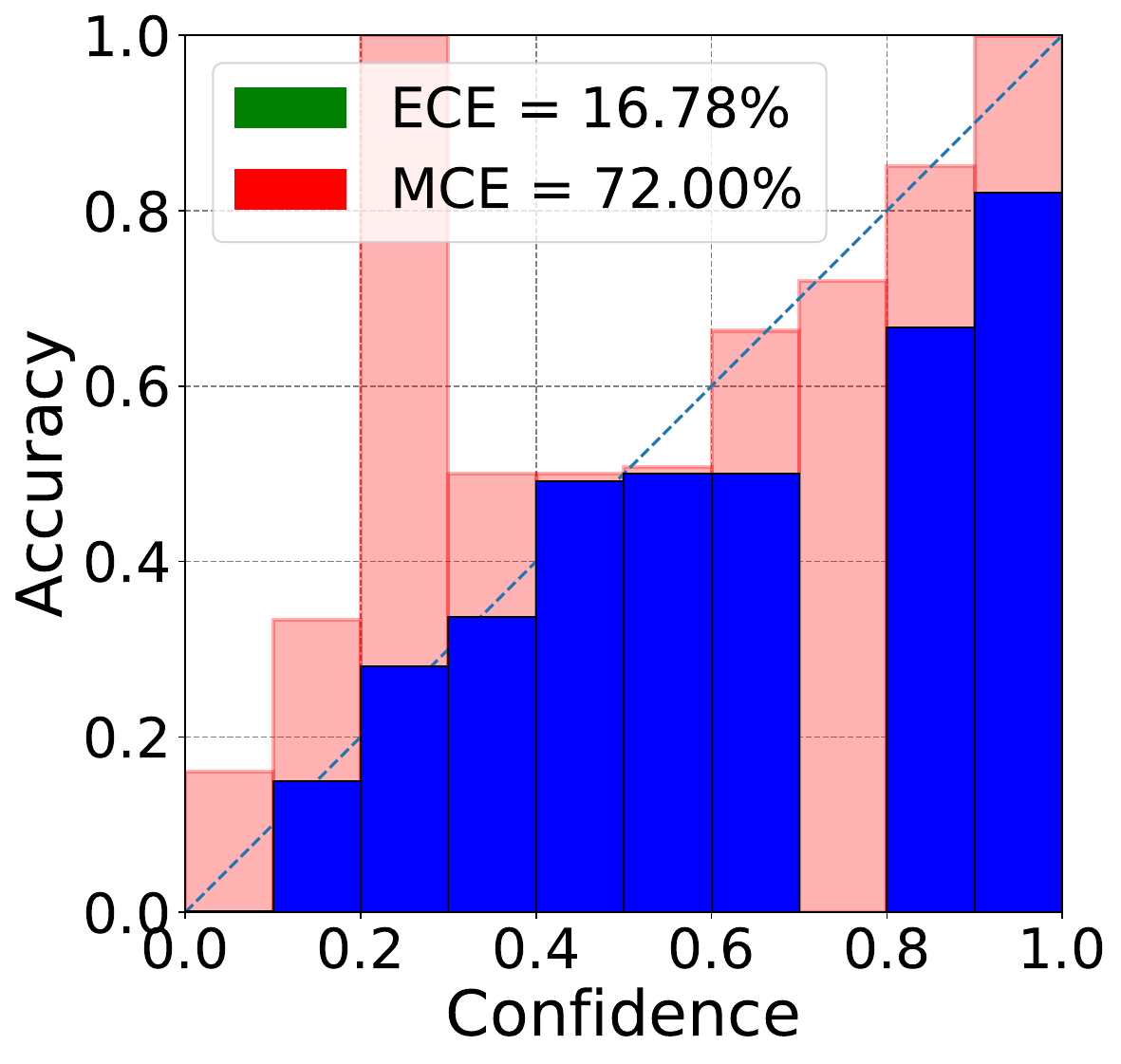}} \hspace{3mm}
    \subfloat[Predictive probability (uncalibrated)]{\includegraphics[width=0.325\textwidth]{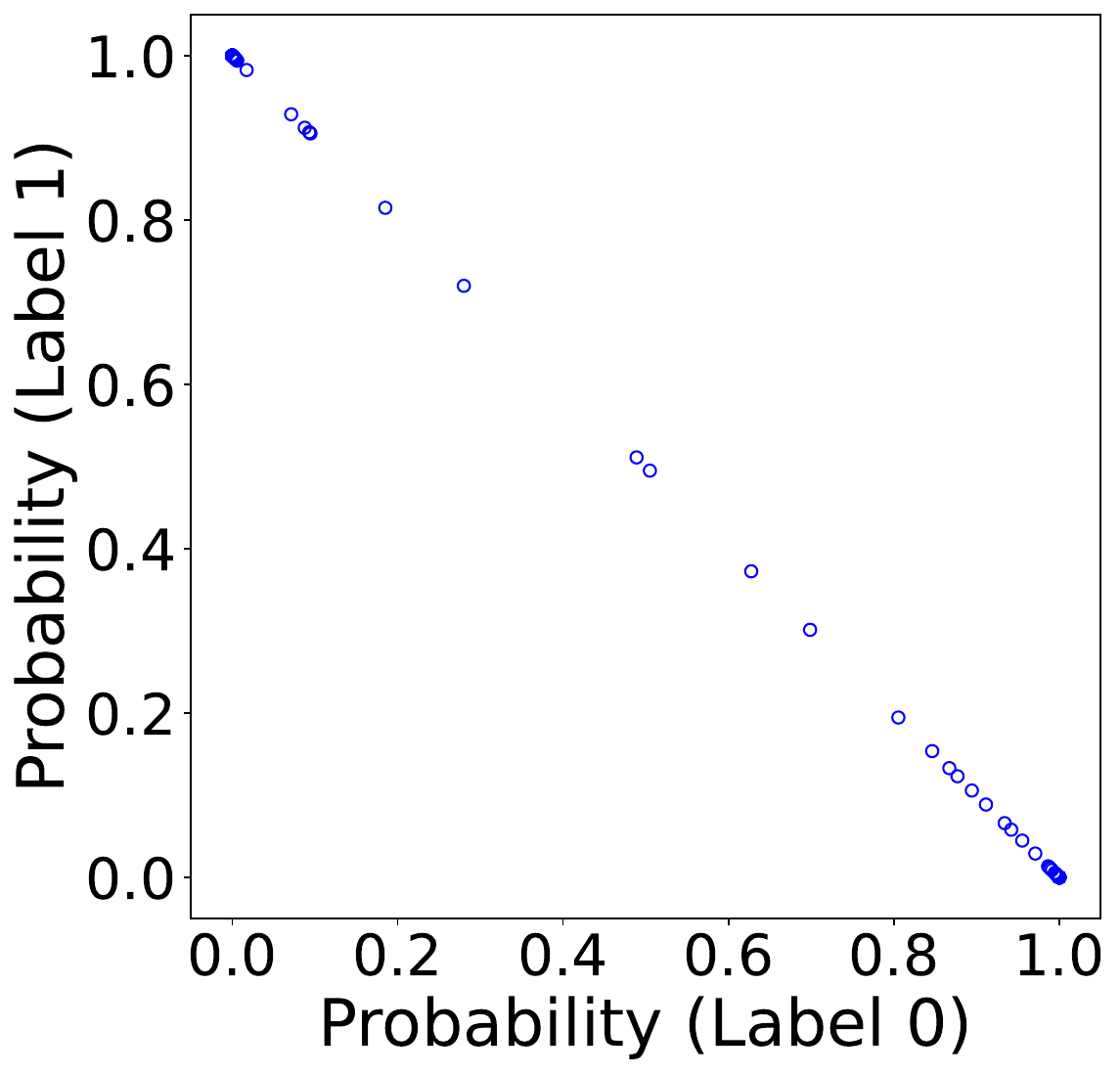}} \\ 
    \subfloat[Reliability diagram (calibrated)]{\includegraphics[width=0.325\textwidth]{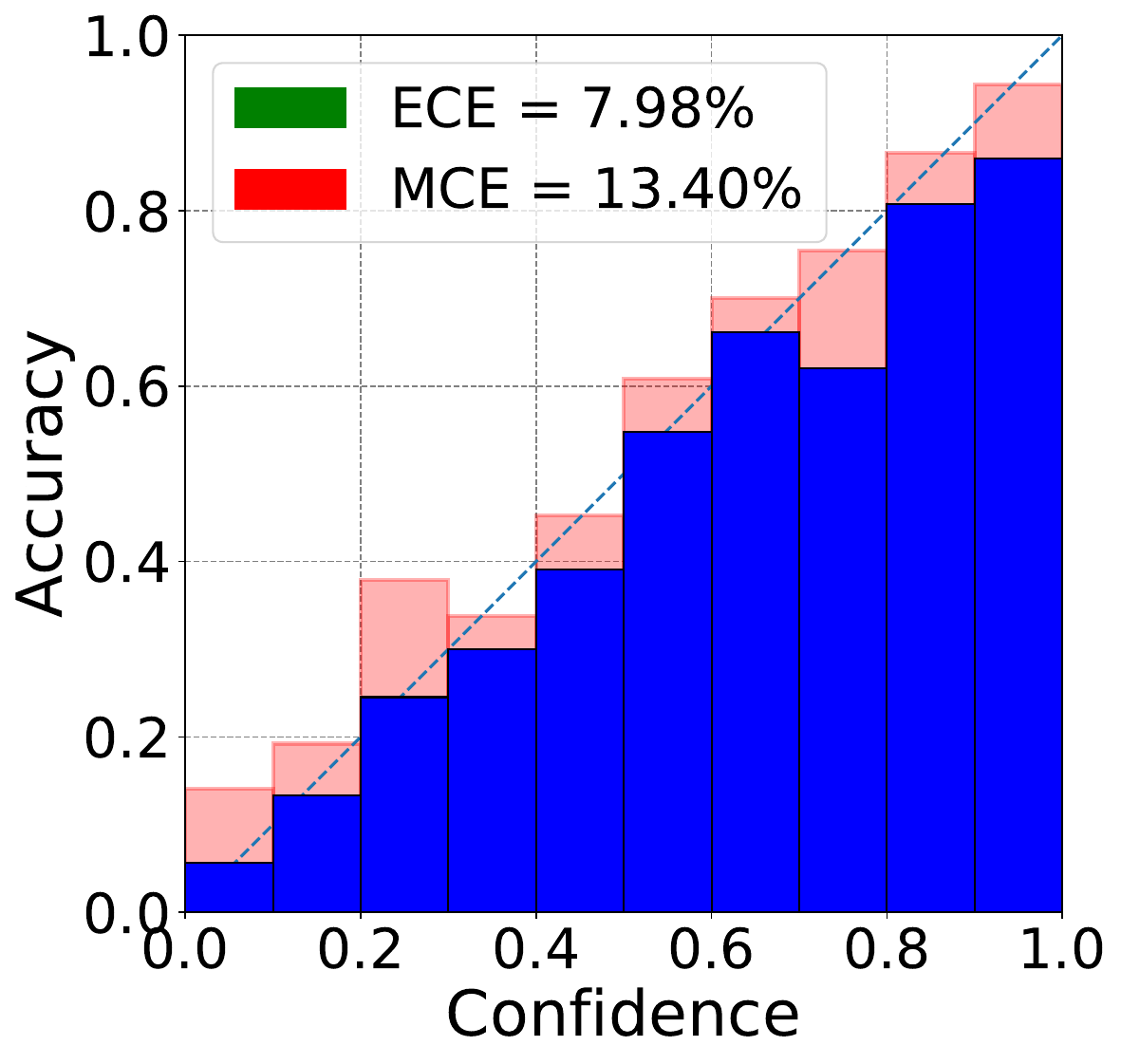}} \hspace{3mm}
    \subfloat[Predictive probability (calibrated)]{\includegraphics[width=0.325\textwidth]{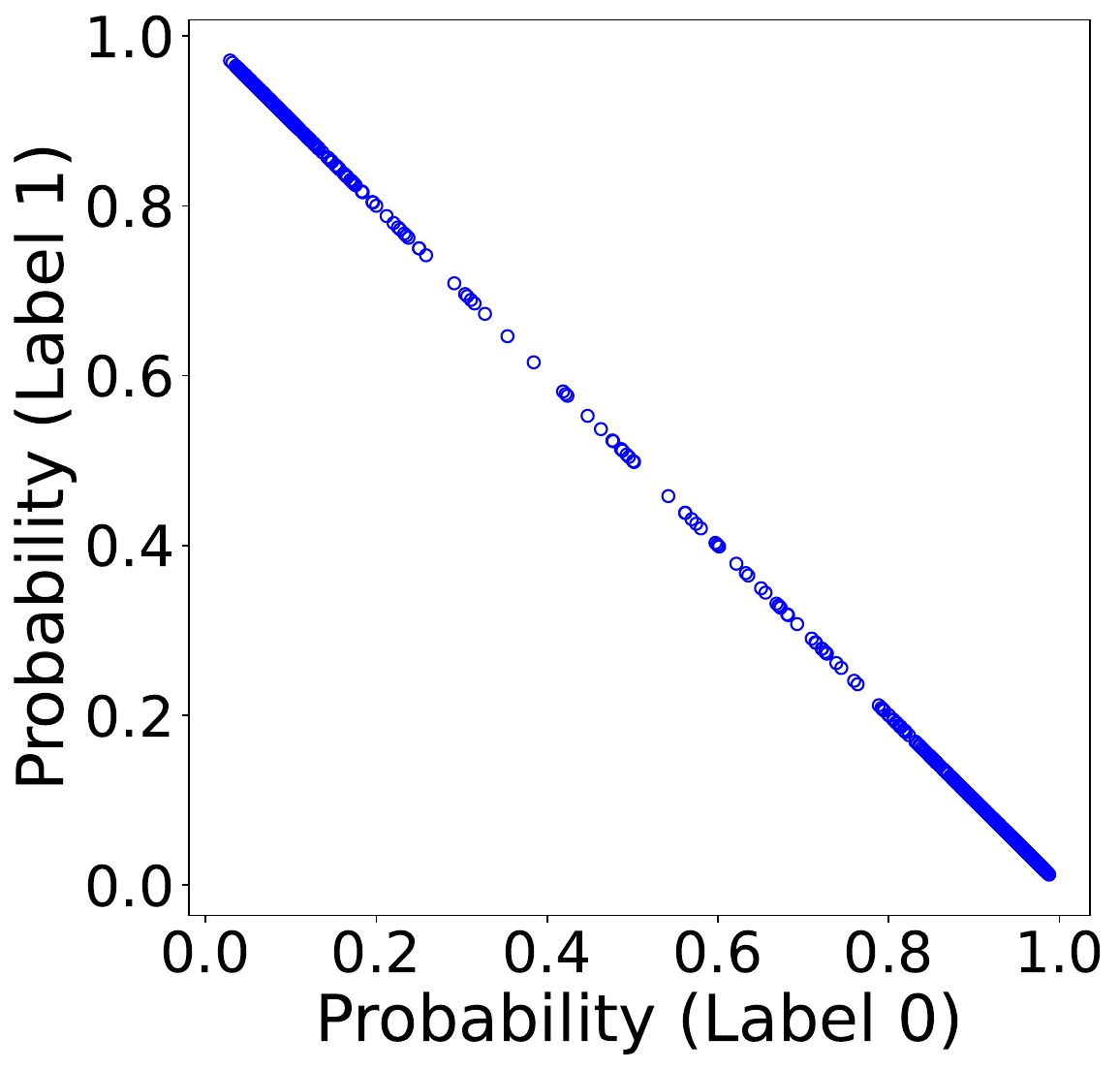}}
    \caption{ Comparison of reliability diagrams and predictive probability between uncalibrated and calibrated binary classification models. (a) and (b) correspond to the uncalibrated model trained with standard cross-entropy loss;  (c) and (d) show the calibrated model trained with focal loss ($\gamma=5$). 
    Both models achieve similar accuracy (83.8\% and 83.4\%, respectively), but differ in calibration. Reliability diagrams (a) and (c) plot predicted confidence (x-axis) versus empirical accuracy (y-axis) using 10 bins. 
    The dashed diagonal represents perfect calibration, while the Expected Calibration Error (ECE) and Maximum Calibration Error (MCE) quantify calibration performance. Histograms (b) and (d) display the distribution of predicted probabilities on 1000 test samples. Pink bars indicate the calibration error (i.e., the absolute difference between average confidence and empirical accuracy) in each bin. 
    }
    \label{fig:calibration}
\end{figure}

Calibrating deep models is a procedure for preventing the model's posterior distribution from being over- or under-confident. Figure~\ref{fig:calibration} gives an illustration of calibrating a binary classification model. It is noted that (1) a highly predictive model can be poorly calibrated, this is exhibited by high calibration errors; (2) The deep models tend to be primarily overconfident, this is shown by a spiking posterior distribution ~\cite{guo2017calibration,wang2021rethinking,mukhoti2020calibrating}. Model miscalibration is caused by many factors such as over-parameterized networks, a lack of appropriate regularization, limited data, imbalanced label distributions, etc. We discuss those factors in details in the Section~\ref{sec:aspects_miscalibration}. 

\subsection{Scope and Focus}
In recent years, a variety of approaches have emerged for model calibration, spanning post-hoc adjustments, regularization techniques, uncertainty estimation, and hybrid methods. Several related surveys have delved into this area~\cite{silva2023classifier}, or those focusing on the highly relevant topic of uncertainty estimation~\cite{silva2023classifier,gawlikowski2021survey,mena2021survey}, which briefly touch upon model calibration. However, there remains a gap in the literature regarding a comprehensive review of recently proposed calibration techniques. Setting itself apart from previous surveys, our research emphasizes several key distinctions:
\begin{itemize}
\item This survey reviews the state-of-the-art calibration methods and focuses mostly on the ones proposed in recent years. Those methods such as kernel-based methods, differentiable calibration proxy, and meta-learning-based approaches. Those are rarely discussed in previous and existing surveys.
\item This survey tries to explain calibration principles of each method via the discussion of the conceptual relationships among over-parameterization, over-fitting, and over-confidence. We systematically categorize those methods into post-hoc, regularization (explicit, implicit and differentiable calibration proxy), uncertainty estimation, and hybrid methods that combines multiple calibration techniques. 
\item This survey also delves into the methodologies for calibrating large pre-trained models, with a specific focus on large language models (LLMs). Notably, the calibration of LLMs for zero-shot inference has garnered escalating interest within AI communities.
\end{itemize}

The rest of this survey is structured as follows. 
\begin{itemize}
\item Section~\ref{sec_pre} provides an introduction to the concept of model calibration, elucidating the factors that contribute to miscalibration. We explore how various aspects such as model complexity, data distribution, training and measurement procedures can impact the calibration of machine learning models.

\item Section~\ref{sec:cal_metric} presents an overview of mainstream calibration metrics commonly used to evaluate the calibration performance of models. We discuss the significance of metrics such as expected calibration error (ECE), reliability diagrams, and other calibration metrics in assessing the alignment between predicted probabilities and true outcomes.

\item Section~\ref{sec:cal_methods} delves into a comprehensive review and classification of recently proposed calibration methods. We categorize these methods based on their underlying principles and techniques, including post-hoc adjustments, regularization approaches, uncertainty estimation techniques, and novel hybrid methods. For each category, we provide in-depth discussions, highlighting their strengths and limitations.

\item Section~\ref{sec:future} discusses potential future directions in the field of model calibration. We identify emerging trends, research gaps, and challenges that warrant further investigation. Additionally, we offer insights into promising avenues for advancing calibration techniques.
\end{itemize}

\section{Preliminaries and Backgrounds}\label{sec_pre}
This section describes the definition of model calibration and the aspects cause miscalibration.
\subsection{Definitions}
In classification tasks, for a given input variable $X$ and a categorical variable $Y \in \{1,2,...,k\}$, assume we have a neural network model $f$ which maps input variable $\bfx $ to a categorical distribution $p=\{p_1,...,p_k\}$ over $k$ classes $\{y_1,...,y_k\}$: $f:D \rightarrow \Delta$, where $\Delta$ is the $k-1$ dimensional standard probability simplex and $\Delta =\{p \in [0, 1]^k |\sum_{i=1}^k p_i=1\}$. \textit{Calibration} measures the degree of the match between predicted probability $p$ and the true correctness likelihood. A model $f$ is perfectly calibrated on if and only if:
\begin{equation}
\mathbb{P}(Y=y_i | f(X)=p) = p_i
\end{equation}
where $p$ is true correctness likelihood. Intuitively, for all input pairs $\{x, y\} \in D$, if model predicts $p_i = 0.8$, we expect that 80\% have $y_i$ as label. 

Instead of using the full probability distribution, the \textit{argmax calibration}~\cite{guo2017calibration,minderer2021revisiting,kumar2018trainable} takes only the maximum probability into consideration:
\begin{equation}
\mathbb{P}(Y =\arg\max(p) | \max(f(X))=p^{*}) = p^{*}
\end{equation}

In reality, it is difficult to obtain \textit{perfect calibration}, any deviation from it represents \textit{miscalibration}.

\subsection{Aspects Impact Model Calibration}
\label{sec:aspects_miscalibration}
\begin{figure}[!htb]
	\centering
	\includegraphics[width=0.8\textwidth]{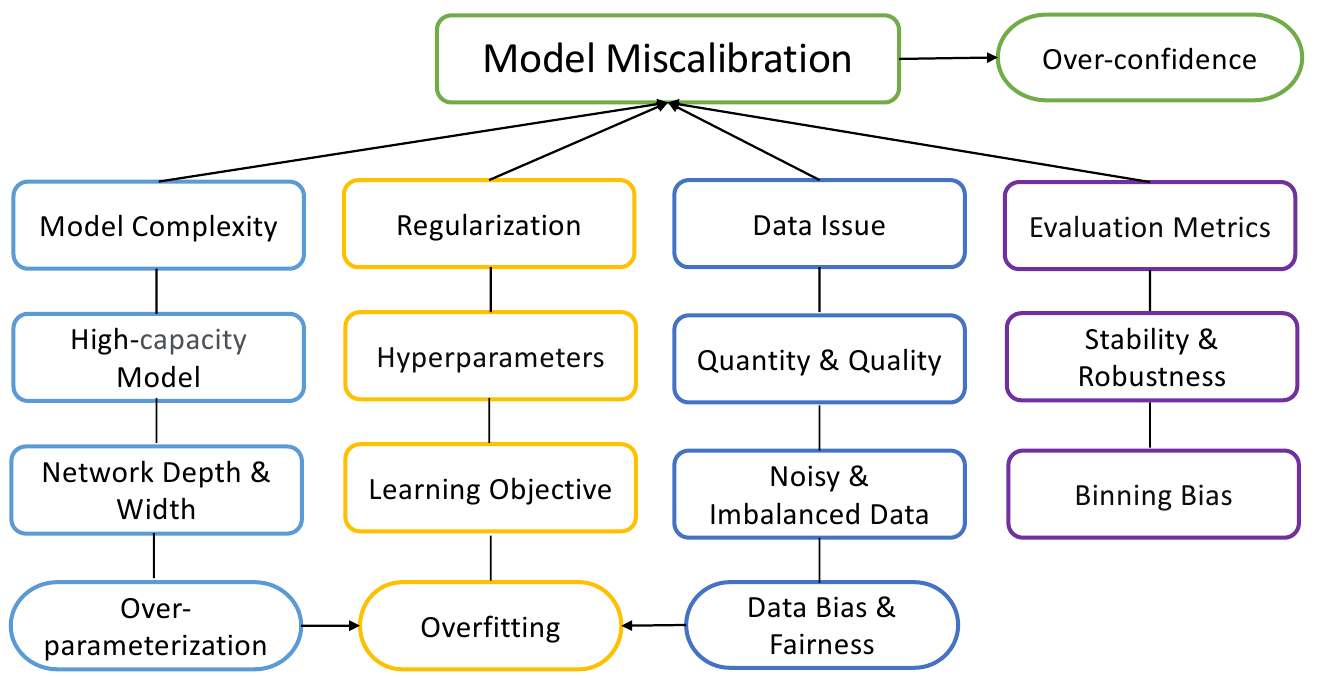} 
	\caption{Illustration depicting factors influencing model calibration. Modern neural networks often exhibit overconfidence, which is closely associated with overfitting during model training. Overfitting is commonly exacerbated by inadequate regularization techniques and the presence of data bias within training datasets, particularly in over-parameterized models.}
\label{fig:aspect_impact_calibration}
\end{figure}

It has been observed that some recent changes in modern neural networks are responsible for model miscalibration~\cite{guo2017calibration,mukhoti2020calibrating,minderer2021revisiting}. The underlying general cause is that modern neural networks' high capacity makes them vulnerable to miscalibration, which is tightly correlated to the concepts of \textit{over-parameter}, \textit{overfitting} and \textit{over-confidence}. Figure~\ref{fig:aspect_impact_calibration} outlines those factors that impact model calibration.

\subsubsection{Model Complexity}
While increasing the depth and width of neural networks helps to obtain highly predictive power, it also negatively increases calibration errors. Empirical evidence has shown that this poor calibration is linked to overfitting on the negative log-likelihood (NLL) \cite{guo2017calibration,mukhoti2020calibrating}. The over-parameterization is one of the main causes of overfitting. Concretely, along with the standard NLL-based model training, when classification error is minimized, keeping training will further push the model to minimize NLL on the training data, i.e., push the predicted softmax probability distribution as close as possible to the ground-truth distribution (which is usually one-hot representation). Model overfitting starts by exhibiting increased test NLL, and then the model becomes overconfident~\cite{guo2017calibration}. This phenomena becomes more severe when training a high capacity model with limited data. 


\subsubsection{Data Issues}
Another important aspects that directly impact model calibration are: data quantity, i.e., the scale of dataset and data quality such as relevance, consistency, completeness, etc. Training high-capacity (over-parameterized) networks with scarce data can easily cause overfitting and have an overconfident model. Data augmentation is an effective way to alleviate this phenomenon and brings implicit calibration effects~\cite{thulasidasan2019mixup,muller2019does}. The recent pretrain-finetune paradigm offers the possibility of reducing overfitting caused by limited data~\cite{desai2020calibration}. Besides having a decent amount of training data, we should also highlight the importance of addressing the following issues:

\begin{itemize}
  \item Data noise: This refers to the presence of erroneous or misleading information in datasets, which can adversely affect model training and performance. Noisy data can arise from various sources such as systematic or measurement errors when pulling historical data from production models, labeling or annotation mistakes caused by human annotators, or outliers ~\cite{brodley1999identifying}. Those factors can lead to biased model estimates, reduced predictive accuracy, and poor generalization to unseen data. This contributes to model miscalibration. Some methods have been proposed to mitigate the impact of data noise on model calibration such as robust loss functions or adversarial training, have shown promise in improving model robustness to noisy data~\cite{hendrycks2018deep,delaney2021uncertainty}. Ensemble methods have also been proven effective in identifying and mitigating the effects of noisy data on model performance~\cite{lakshminarayanan2017simple,malinin2019ensemble}..
  
  \item Data imbalance: It can significantly impact model calibration. When a dataset is imbalanced, meaning that one class or outcome is significantly more prevalent than others, it can lead to biased predictions and poor calibration. This is because the model may become overly confident in predicting the majority class while being less certain about the minority classes. Focal loss ~\cite{lin2017focal,mukhoti2020calibrating} has shown its effectiveness in calibrating models that trained with imbalanced datasets~\cite{wang2022calibrating}. 
  
  \item Data biases and fairness: Extensive studies have highlighted the presence of biases in datasets. Biases arises due to historical disparities, societal prejudices, or systemic inequalities. For instance, facial recognition datasets often exhibit racial and gender biases~\cite{buolamwini2018gender}. Mitigating bias and ensuring fairness are imperative steps towards building models that exhibit better calibration and reliability~\cite{pleiss2017fairness,calegari2023assessing}. This involves a multifaceted approach that begins with critically examining the data collection process to identify and rectify sources of bias. Techniques such as data augmentation~\cite{muller2019does,zhang2018mixup}, bias mitigation algorithms~\cite{kamiran2012data}, and fairness-aware model training~\cite{zemel2013learning} can help mitigate biases during the model development stage. 
\end{itemize}

\subsubsection{Learning Objective and Regularization}

The choice of loss function and optimization objective can significantly impact the calibration quality of machine learning models. For instance, cross-entropy loss is commonly used for classification tasks, as it directly optimizes for the likelihood of correct class predictions. However, optimizing solely for accuracy may lead to poorly calibrated probability estimates, especially in situations where class distributions are imbalanced or ambiguous. Focal loss~\cite{lin2017focal,mukhoti2020calibrating}, as alternative loss,  has recently demonstrated promising performance in calibrating deep models.

Regularization can effectively prevent overfitting when model capacity increases. Recent trends suggest that explicit L2 regularization may not be necessary to achieve a highly accurate model when applying batch normalization~\cite{ioffe2015batch} or dropout~\cite{srivastava2014dropout}, but Guo et al.~\cite{guo2017calibration} empirically demonstrated model tends to be less calibrated without using L2 regularization. There have been more recent regularization techniques ~\cite{mukhoti2020calibrating,bohdal2021meta,kumar2018trainable,pereyra2017regularizing} been proposed to improve model calibration. We will discuss those methods in more details in the Section~\ref{sec:cal_methods}.

\subsubsection{Bias in Evaluation Metrics}
Another important factor that does not directly cause model calibration, but hurts accurate assessment of model calibration is the introduced bias in evaluation metrics. For instance, the bias that introduced by binning mechanisms in the widely used Expected Calibration Error (ECE) has been identified in ~\cite{nixon2019measuring}. By binning continuous variables, fine-grained details in the data may be lost, leading to an oversimplified representation of the underlying relationships. This loss of information can result in calibration bias, as the model's predicted probabilities may not accurately capture the nuances of the data distribution.

To mitigate calibration bias introduced by binning mechanisms , it is important to carefully consider the choice of bin boundaries and the impact of discretization on the model's calibration performance. There have been some techniques proposed to address this issues such as Adaptive ECE~\cite{nixon2019measuring}, Smooth ECE~\cite{wang2023meta}. We will further discuss in the Section~\ref{sec:cal_methods}.

\section{Calibration Measurements}
\label{sec:cal_metric}
Exact calibration measurement with finite data samples is impossible given that the confidence $p$ is a continuous variable~\cite{guo2017calibration}. There are some popular metrics that approximate model calibration error by grouping $N$ predictions into $M$ interval bins $\{b_1, b_2,...,b_M\}$.

\subsection{Expected Calibration Error (ECE)}

ECE~\cite{NaeiniETAL:15} is a scalar summary statistic of calibration. It quantifies the discrepancy between predicted probabilities and observed frequencies by dividing the prediction space into several bins based on the predicted probabilities. Then, within each bin, it calculates the average absolute difference between the predicted probabilities and the empirical frequencies of the events occurring. The ECE is computed as the weighted average of these differences across all bins. Mathematically, ECE can be expressed as:
\begin{equation}
\textsc{ECE} =\frac{1}{N} \sum_{m=1}^{M} \left | b_m \right | |\textrm{acc}(b_m) - \textrm{conf}(b_m)|
\end{equation}
where $N$ is the total number of samples. $\left | b_m \right |$ is the number of samples in bin $b_m$, and 
\begin{equation}
\textrm{acc}(b_m)=\frac{1}{\left|b_m\right|}\sum_{i \in B_m}\mathbf{1}(\hat{y}_i=y_i),~
\textrm{conf}(b_m)=\frac{1}{\left|b_m\right|}\sum_{i \in B_m}p_i.
\end{equation}

\subsection{Maximum Calibration Error (MCE)}
MCE~\cite{NaeiniETAL:15}, MCE focuses on identifying the largest discrepancy between predicted probabilities and empirical accuracies within any individual bin,
\begin{equation}
\textsc{MCE} = \max_{m\in \{1, \dots, M\}} | \textrm{acc}(b_m) - \textrm{conf}(b_m)|.
\end{equation}
and is particularly important in high-risk applications where reliable confidence measures are absolutely necessary.  While MCE highlights the most substantial calibration error observed within the prediction space, it may not offer as comprehensive an assessment of overall calibration performance as ECE does. ECE considers the average discrepancy across all bins, providing a more holistic evaluation of calibration quality.

\subsection{Brier Score}
The Brier Score~\cite{brier1950verification} is a proper scoring rule that measures both discrimination and calibration of probabilistic predictions. For binary classification, it computes the mean squared error between predicted probabilities and actual outcomes. The metric can be formulated as
\begin{equation}
\textsc{BS} = \frac{1}{N} \sum_{i=1}^{N} (f_i - y_i)^2
\end{equation}
where $f_i$ represents the predicted probability for the positive class and $y_i \in {0,1}$ is the true binary label. For multi-class problems with $K$ classes, the Brier Score can be extended as
\begin{equation}
\textsc{BS} = \frac{1}{N} \sum_{i=1}^{N} \sum_{c=1}^{K} (f_{i,c} - y_{i,c})^2
\end{equation}
where $f_{i,c}$ is the predicted probability for class $c$ and $y_{i,c}$ is the binary indicator for the true class.

The Brier Score ranges from 0 to 1, where 0 indicates perfect calibration and discrimination. It can be decomposed into reliability (calibration), resolution, and uncertainty components, providing insights into different aspects of predictive performance.

\subsection{Classwise ECE (CECE)}
Classwise ECE~\cite{kull2019beyond} can be seen as the macro-averaged ECE. It extends the bin-based ECE to measure calibration across all the possible $K$ classes. In practice, predictions are binned separately for each class, and the calibration error is computed at the level of individual class-bins and then averaged. The metric can be formulated as
\begin{equation}
\textsc{CECE} = \sum_{m=1}^{M} \sum_{c=1}^{K} \frac{|b_{m, c}|}{N K} |\textrm{acc}_c(b_{m, c}) - \textrm{conf}_c(b_{m, c})|
\end{equation}
where $b_{m, c}$ represents a single bin for class $c$. In this formulation, $\textrm{acc}_c(b_{m, c})$ represents average binary accuracy for class $c$ over bin $b_{m, c}$ and $\textrm{conf}_c(b_{m, c})$ represents average confidence for class $c$ over bin $b_{m, c}$. 

\subsection{Adaptive ECE (AECE)}
The binning mechanism in the aforementioned metrics can introduce bias; the pre-defined bin size determines the number of samples in each bin. Adaptive ECE~\cite{nixon2019measuring} introduces a new binning strategy to use an adaptive scheme that spaces the bin intervals to ensure each bin hosts an equal number of samples.
\begin{equation}
\textsc{AECE} = \sum_{r=1}^{R} \sum_{c=1}^{K} \frac{1}{RK} |\textrm{acc}_c(b_{n, c}) - \textrm{conf}_c(b_{n, c})|
\end{equation}
Where $r \in [1,R]$ is defined by the $[N/R]$-th index of the sorted and threshold predictions.

For a perfectly calibrated classifier, those calibration errors should equal $0$.

\subsection{Reliability Diagram}
Besides the metrics that provide a scalar summary on calibration, reliability diagram (also referred to as a calibration plot or calibration curve \ref{fig:calibration}) visualizes whether a model is over- or under-confident on bins by grouping predictions into bins according to their prediction probability. This visual representation enables a detailed examination of how well the predicted probabilities align with the observed frequencies of outcomes, providing insights into the reliability of the model's confidence estimates.The diagonal line in Figure 1 presents perfect calibration: 
\begin{equation}
\textrm{acc}(b_m)=\textrm{conf}(b_m), \forall m
\end{equation}
The red bar presents the gap to perfect calibration. Deviations from the perfect calibration line indicate calibration errors. Points above the diagonal suggest that the model's predicted probabilities are overly confident, while points below the diagonal indicate underconfidence. The magnitude and direction of these deviations provide valuable insights into the systematic biases present in the model's probabilistic predictions.

\section{Calibration Methods}
\label{sec:cal_methods}
In this section, we categorize the state-of-the-art calibration methods into post-hoc methods, regularization methods and uncertainty estimation methods. Besides, we discuss hybrid methods that combine different calibration methods. Figure \ref{fig:categorization} summarizes and categories those methods.

\begin{figure}
\fontsize{6.5pt}{7pt}\selectfont
\begin{forest}
    for tree={
        draw, semithick, rounded corners, drop shadow,
        text centered,
        edge={draw, semithick},
        anchor=east,
        grow=east,
        forked edge,
        s sep=2mm,
        l sep=2mm,
        fork sep=2mm
    }
[Calibration, top color=red!20, bottom color=red!40
    [Large Language Models, top color=blue!20, bottom color=blue!40,
        for descendants={fill=blue!10}
        [Contextual Calibration~\cite{zhao2021calibrate}]
        [Prototypical Calibration~\cite{han2022prototypical}]
        [Domain-Context Calibration~\cite{fei2023mitigating}]
        [Batch Calibration~\cite{zhou2023batch}]
        [PriDe~\cite{zheng2023large}]
        [Thermometer~\cite{shen2024thermometer}]
    [Atomic Calibration\cite{zhang2024atomic}]
    ]
    [Hybrid Methods, top color=green!20, bottom color=green!40,
        for descendants={fill=green!10}
        [LS + Mixup~\cite{thulasidasan2019mixup}]
        [Ensemble TS (ETS)~\cite{zhang2020mix}]
        [TS + MC Dropout~\cite{laves2019well}]
        [Conformal Prediction + MC Dropout~\cite{cocheteux2025uncertainty}]
    ]
    [Uncertainty Estimation, top color=purple!20, bottom color=purple!40,
        for descendants={fill=purple!10}
        [Sample-weighted Gradient
        [BSCE‑GRA~\cite{lin2025uncertainty}]
        ]
        [Bayesian Methods
            [Finite-State Probabilistic RNNs~\cite{wang2020uncertainty}]
            [MC Dropout~\cite{gal2016dropout}]
            [Bayesian RNNs~\cite{fortunato2017bayesian}]
            [BNNs~\cite{blundell2015weight}]
        ]
        [Ensemble Methods
            [Deep Ensembles~\cite{lakshminarayanan2017simple}]
            [Adaptive Calibrator Ensemble~\cite{zou2023adaptive}]
        ]
    ]
    [Regularization Methods, top color=orange!20, bottom color=orange!40,
        for descendants={fill=orange!10}
        [Explicit Regularization
            [Huber Loss~\cite{patra2023calibrating}]
            [DCA~\cite{liang2020improved}]
            [Entropy Regularization~\cite{pereyra2017regularizing}]
        ]
        [Implicit Regularization
            [Focal Loss~\cite{lin2017focal}]
            [FLSD~\cite{mukhoti2020calibrating}]
            [Label Smoothing (LS)~\cite{muller2019does}]
            [Mixup~\cite{thulasidasan2019mixup}]
        ]
        [Differentiable Proxy
            [MMCE~\cite{kumar2018trainable}]
            [Meta-Calibration~\cite{bohdal2021meta}]
            [AvUC~\cite{krishnan2020improving}]
            [MDCA~\cite{hebbalaguppe2022stitch}]
        ]
    ]
    [Post-hoc Methods, top color=gray!20, bottom color=gray!40,
        for descendants={fill=gray!10}
        [Top-versus-All (TvA)~\cite{le2024confidence}]
        [Density‑Aware Calibration (DAC)~\cite{tomani2023beyond}]
        [One-vs-All Surrogate Loss~\cite{verma2022calibrated}]
        [Local TS (LTS)~\cite{ding2021local}]
        [Dirichlet Calibration~\cite{kull2019beyond}]
        [Bin-wise TS (BTS)~\cite{ji2019bin}]
        [Attended TS (ATS)~\cite{mozafari2018attended}]
        [Temperature Scaling (TS)~\cite{guo2017calibration}]
    ]
]
\end{forest}
\caption{Categorization of calibration methods and representative approaches.}
\label{fig:categorization}
\end{figure}

\subsection{Post-hoc Methods}
Post-hoc calibration methods aim to calibrate a model after training. Those include non-parametric calibration histogram binning\cite{zadrozny2001obtaining}, isotonic regression~\cite{zadrozny2002transforming} and parametric methods such as Bayesian binning into quantiles (BBQ) and Platt scaling~\cite{platt1999probabilistic}. Out of them, Platt scaling~\cite{platt1999probabilistic} based approaches are more popular due to their low complexity and efficiency. This includes \textit{Temperature Scaling (TS)} and its extensions such as attended TS~\cite{mozafari2018attended}, \textit{Dirichlet calibration}~\cite{kull2019beyond}, \textit{Bin-wise TS (BTS)}~\cite{ji2019bin} and  \textit{Local TS (LTS)}~\cite{ding2021local}.

\subsubsection{Temperature Scaling}
\textit{Temperature scaling (TS)} is a single-parameter extension of Platt scaling~\cite{platt1999probabilistic} and the most recent addition to the offering of post-hoc methods. It uses a temperature parameter $\tau$ to calibrate the softmax probability:
\begin{equation}
p_i= \frac{\exp (g_i/\tau)}{\sum_{j=1}^{k} \exp (g_j/\tau)}, ~~i \in [1\dots k].
\label{eq:diffrentiable_approxmiation}
\end{equation}
where $g_i$ is the logit and $\tau>0$ for all classes is used as a scaling factor to soften model predicted probability, it controls the model's confidence by adjusting the sharpness of distribution so that the model prediction is not too certain (overconfident) or too uncertain (underconfident). The optimal temperature value is obtained by minimizing negative log likelihood loss (NLL) on the validation dataset.:
\begin{equation}
\tau^{\ast}=\arg \min_{\tau}\left (-\sum_{i=1}^N\log(\textsc{softmax}(g_i,\tau))  \right )
\label{eq:diffrentiable_approxmiation}
\end{equation}
TS simplifies matrix (vector) scaling~\cite{guo2017calibration} where class-wise $\tau$ is considered as a single parameter, and offers good calibration while maintaining minimum computational complexity~\cite{guo2017calibration,minderer2021revisiting}.

\subsubsection{Temperature Scaling Extensions}
The goal of post-hoc calibration on a validation dataset is to learn a \textit{calibration map} (also known as the canonical calibration function of probability~\cite{vaicenavicius2019evaluating}) which transforms uncalibrated probabilities into calibrated ones. Many TS extensions aim to find a proper calibration map.

Kull et al.~\citeyear{kull2019beyond} proposed \textit{Dirichlet calibration} which assumes probability distributions are parameterized Dirichlet distributions:
\begin{equation}
p(x)|y = j \sim Dirichlet(\alpha^j)
\end{equation}
where $\alpha^j=\{\alpha^j_1, \alpha^j_2,...,\alpha^j_k\}$ are Dirichlet parameters for the $j$-th class among $k$ total classes. The proposed Dirichlet calibration map family coincides with the Beta calibration family~\cite{pmlr-v54-kull17a}. Besides, it provides uniqueness and interpretability as compared to generic canonical parametrization.

Mozafari et al.~\citeyear{mozafari2018attended} suggested that TS has difficulties in finding optimal $\tau$ when the validation set has a limited number of samples. They proposed \textit{attended temperature scaling (ATS)} to alleviate this issue by increasing the number of samples in validation set. The key idea is to gather samples from each class distribution. Let's assume $p(y|x)$ is the predicted probability. ATS first divides the validation set into $K$ subsets with $p(y=k|x), k \in [1, K]$, which allows to add more $y \neq k$ samples using the Bayesian Theorem~\cite{jin2017introspective} as the selection criterion:
\begin{equation}
p(x, y=k) = \frac{p(y=k|x)}{p(y\neq k|x)}p(x, y \neq k)
\end{equation}
It indicates that ATS selects the samples with $y\neq k$ which are more probable to belong to $p(x, y=k)$.
 
\textit{Bin-wise TS (BTS)}~\cite{ji2019bin} was proposed to extend TS to multiple equal size bins by using the confidence interval-based binning method. Together with data augmentation, BTS showed superior performance as compared to TS.
\textit{Local TS (LTS)}~\cite{ding2021local} extends TS to multi-label semantic segmentation and makes it adaptive to local image changes. A local scaling factor is learned for each pixel or voxel.
\begin{equation}
\mathbb{Q}(x, \tau_i(x)) = max_{l \in L}\textsc{Softmax}(\frac{g_i(x)}{\tau_i(x)})^{(l)}
\end{equation}
where $l$ is the class index, $g_i(x)$ is the logit of input at location $x$, and The $\tau_i(x)$ is the location-dependent temperature.

Verma and Nalisnick~\citeyear{verma2022calibrated} revisit the ``learning to defer'' (L2D) framework, which allows a model to defer decisions to a human expert when appropriate. They identify that the prevalent multiclass L2D approach by ~\cite{pmlr-v119-mozannar20b}, which uses a single softmax over classes plus a reject option, is intrinsically miscalibrated with respect to predicting expert correctness—its softmax parameterization can yield probabilities summing to more than one and does not reflect true likelihoods of expert success. To correct this, the authors propose an alternate scheme using one‑vs‑all (OvA) classifiers: each class (including expert correctness) is modeled with its own sigmoid output, forming a vector of calibrated probabilities. They derive a consistent surrogate loss (denoted $\psi_{\text{OvA}}$) that reduces to multiple binary classification problems. This construction guarantees well-calibrated probabilities while maintaining consistency with the 0–1 L2D objective. Empirical evaluation across tasks such as hate speech detection, galaxy classification, and skin lesion diagnosis shows that the OvA‑based model attains calibration without sacrificing—often improving—accuracy compared to the softmax baseline.

Le Coz \emph{et~al.}~\citeyear{le2024confidence} proposed the \textit{Top-versus-All (TvA)} post‑hoc calibration method, which reformulates multiclass confidence calibration into a single binary task. They construct a surrogate classifier by taking the maximum softmax score and assigning a binary label of correctness. This setup allows applying standard binary calibration tools such as Temperature Scaling~\cite{guo2017calibration}, Histogram Binning~\cite{zadrozny2001obtaining} and Dirichlet Calibration~\cite{kull2019beyond} more effectively, avoiding issues of multiclass imbalance and complexity. In the case of scaling methods, they replace the usual cross‑entropy with binary cross‑entropy, creating stronger gradients that penalize incorrect overconfidence. They also introduce simple $L_2$ regularization for Vector/Dirichlet Scaling to reduce overfitting in large‑class regimes. Extensive experiments on image and text–based models (including PLMs and LLMs) demonstrate that TvA improves calibration metrics (like ECE), preserves predicted labels, and is easy to plug into existing pipelines.

Tomani\emph{et~al.}~\citeyear{tomani2023beyond} introduced \textit{Density‑Aware Calibration (DAC)}, a post‑hoc method designed to improve calibration under domain shifts and OOD scenarios. DAC uses k‑nearest neighbor density estimates in hidden-layer feature space to measure how “in‑distribution” a test sample is, then rescales its logits based on this density before applying any standard post‑hoc calibrator $h$ (e.g., TS, Dirichlet/Vector Scaling, splines). This preserves in-domain accuracy while significantly reducing Expected Calibration Error (ECE) under corruptions and OOD. Experiments on CIFAR‑C and ImageNet‑C with CNNs and Transformers show that DAC enhances robustness and calibration without retraining the original model.

\subsubsection{Practical Considerations} 
Despite using only a single global hyperparameter, TS remains widely adopted due to its simplicity, effectiveness, and ability to preserve accuracy~\cite{zhang2020mix}. Post-hoc methods are appealing because they decouple calibration from training, require minimal computational overhead, and perform well even with limited validation data. However, their expressiveness is often constrained. While extensions like Dirichlet~\cite{kull2019beyond}, Bin-wise~\cite{ji2019bin}, and Local TS~\cite{ding2021local} offer more flexibility, they introduce complexity and may require domain-specific tuning. Some methods address data sparsity directly, such as Attended TS~\cite{mozafari2018attended}, which enriches calibration samples through class-aware sampling.

Recent advances extend post-hoc calibration to new settings. Verma and Nalisnick~\citeyear{verma2022calibrated} improve deferral calibration using one-vs-all classifiers, while Le Coz et al.~\citeyear{le2024confidence} reduce multiclass calibration to a binary task via Top-versus-All scoring. Tomani et al.~\citeyear{tomani2023beyond} introduce density-aware scaling to improve calibration under distribution shift. Nonetheless, post-hoc methods depend heavily on the quality and representativeness of validation data. Their limited capacity to approximate complex calibration maps may reduce effectiveness under domain shift or when applied to more nuanced prediction tasks~\cite{zhang2020mix}.

\subsection{Regularization Method}
Regularization is important to prevent neural network models from overfitting. In this section, we discuss some representative work in this direction that either explicitly or implicitly regularizes modern neural networks to have better calibration.

\subsubsection{Explicit Regularization}
The typical (explicit) way to add regularization term (or penalty) $L_{reg}$ to standard loss objective (e.g., negative log likelihood):
\begin{equation}
\mathcal{L}(\theta)= -\sum\log p(y|x) + \alpha L_{reg}.
\end{equation}
where $\alpha$ controls the importance of penalty in regularizing weight $\theta$. \textit{L2 regularization} has been widely used in training modern neural networks and showed its effectiveness in model calibration~]\cite{guo2017calibration}. \textit{Entropy regularization}~\cite{pereyra2017regularizing}:
\begin{equation}
\mathcal{L}(\theta) = -\sum \log p(y \mid x) - \alpha H(p(y \mid x))
\end{equation}
directly penalizes predictive distributions with low entropy, discouraging overly confident (peaked) outputs and promoting better generalization. 

Recently, \cite{liu2022devil} analyzed popular calibration methods such as label smoothing~\cite{muller2019does}, focal loss~\cite{lin2017focal}, and confidence penalties~\cite{pereyra2017regularizing} from a constrained optimization perspective. They showed that these methods can be interpreted as approximations of a linear penalty. Based on this analysis, the authors proposed adding a margin $m_a$ to the learning objective:
\begin{equation}
\min \mathcal{L}_{CE} + \alpha \sum_k \max\bigl(0, \max_j(g_j^{(k)}) - g_{y_k}^{(k)} - m_a\bigr)
\label{eq:margin_based_LS}
\end{equation}
where $g_j^{(k)}$ denotes the logit for class $j$ for input $x_k$, and $g_{y_k}^{(k)}$ is the logit corresponding to the ground-truth class $y_k$.

Liang et al.,~\citeyear{liang2020improved} proposed to directly add the difference between confidence and accuracy (DCA) to standard cross-entropy as a penalty term to improve calibration. \cite{patra2023calibrating} further extend this non-differentiable term to a differentiable version using Huber loss, which is used as an explicit regularization. They also proposed to use a hyper-parameter to control the transition of explicit L1 and L2 regularizer. The Huber loss is then combined with focal loss which provides implicit regularization effects.

\subsubsection{Implicit Regularization: Focal Loss and Its Extensions}
\textit{Focal loss}~\cite{lin2017focal} was originally proposed to alleviate the class imbalance issue in object detection:$
\mathcal{L}_f = -\sum_{i=1}^N(1-p_{i})^{\gamma}\log p_{i}
$
where $\gamma$ is a hyperparameter.
It has been recently shown that focal loss can be interpreted as a trade-off between minimizing Kullback–Leibler (KL) divergence and maximizing the entropy, depending on $\gamma$~\cite{mukhoti2020calibrating}:
\begin{equation}
\mathcal{L}_f \geq \textsc{KL}(q \parallel p) + \mathbb{H}(q) - \gamma \mathbb{H}(p)
\end{equation}
The first term pushes the model to learn a probability $p$ to have a high value (confident, as close as possible to ground-truth label distribution, which is usually one-hot representation). The second term is constant. The last term regularizes probability to not be too high (overconfident).

Mukhoti et al.~\citeyear{mukhoti2020calibrating} empirically observed that $\gamma$ plays an important role in implicitly regularizing entropy and weights. However, finding an appropriate $\gamma$ is challenging for all samples in the datasets. Thus, they proposed \textit{sample-dependent scheduled $\gamma$ (FLSD)} based on the Lambert-W function~\cite{corless1996lambertw}. They have shown that scheduling $\gamma$ values according to different confidence ranges helps to improve model calibration on both in-domain and out-of-domain (OOD) data.

\subsubsection{Implicit Regularization: Differentiable Calibration Proxy}
Recall that the aforementioned methods use a penalty term (either explicitly or implicitly) to improve model calibration on dataset $D$. There is a rising direction that directly optimizes objective function by using calibration errors as differentiable proxy ~\cite{kumar2018trainable,bohdal2021meta} to standard loss:
\begin{equation}
\arg\min_{\theta}L_{standard}(D, \theta) + L_{calibration}(D,\theta)
\end{equation}
The focus of this line of work is to find differentiable approximations to calibration errors.

Kumar et al. ~\citeyear{kumar2018trainable} proposed a kernel-based approach to explicitly calibrate models in training phrase known as\textit{Maximum Mean Calibration Error (MMCE)}, which is differentiable and can be optimized using batch stochastic gradient algorithms. They cast the calibration error to be differentiable by defining an integral probability measure over functions from a reproducing kernel Hilbert space (RKHS) $\mathcal{H}$ induced by a universal kernel $k(\cdot, \cdot)$ and cannonical feature map $\phi:[0,1] \rightarrow \mathcal{H}$:
\begin{equation}
\textsc{MMCE}(P(r, c)) = \left \| E_{(r, c)\sim P}[(c-r)\phi (r)] \right \|_{\mathcal{H}}
\end{equation}
where $r,c$ represent confidence and correctness scores, respectively, and $P(r, c)$ denotes the distribution over $r, c$ of the predicted probability $P(y|x)$.
 
An approach that combines meta-learning and a differentiable calibration proxy was proposed by~\cite{bohdal2021meta}. The authors developed a \textit{differentiable ECE (DECE)} and used it as learning objective for a meta network. The meta network takes representations from original backbone network and outputs a unit-wise L2 weight decay coefficient $\omega$ for backbone network. The DECE is optimized against calibration metrics with validation set but attached to standard cross-entropy (CE) loss.
\begin{align}
\omega^{*} &= \arg\min_\omega \mathcal{L}_{DECE}^{val}(f_c^{*}\circ f_{\theta}^{*}(\omega))\\
f_c^{*}, f_{\theta}^{*}(\omega) &= \arg\min_{f_c, f_{\theta}(\omega)} (\mathcal{L}_{CE}^{train}(f_c\circ f_{\theta}(\omega) + \omega \left \| f_c \right \|^2)
\end{align}
where $f_c$ is classification layer and $f_{\theta}$ is the feature extractor.

Mukhoti and Gal et al.~\citeyear{mukhoti2021deep} proposed patch accuracy versus patch uncertainty (PAvPU) for evaluating uncertain estimates for each image on segmentation task; Motivated by this idea,  ~\cite{krishnan2020improving} extended it to a differentiable proxy, accuracy versus uncertainty calibration (AvUC), which approximates to the accuracy versus uncertainty (AvU):
\begin{equation}
AvU = \frac{n_{AC} + n_{IU}}{n_{AC}+n_{AU}+n_{IC}+n_{IU}}
\end{equation}
The $n_{AC}$, $n_{AU}$, $n_{IC}$, $n_{IU}$ represent the number of samples in the categories of accurate and certain (AC), accurate and uncertain (AU) and certain (IC), inaccurate an uncertain (IU). The AvUC loss is 0 only when all the accurate predictions are certain and inaccurate predictions are uncertain:
\begin{equation}
\mathcal{L}_{AvUC} = -\log(\frac{n_{AC} + n_{IU}}{n_{AC}+n_{AU}+n_{IC}+n_{IU}})
\end{equation}
where $n_{AC}$, $n_{AU}$, $n_{IC}$, $n_{IU}$ are approximated by individual proxy functions. This loss can also be added as regularizer to standard learning objective.

Similarly, Multi-class Difference in Confidence and Accuracy (MDCA)~\cite{hebbalaguppe2022stitch} was proposed to improve model calibration. MDCA is a differentiable calibration loss that does not rely on binning mechanisms, and its formulation is inspired by Static Calibration Error (SCE)~\cite{nixon2019measuring}. The authors demonstrated that combining MDCA with focal loss results in the lowest calibration error among the evaluated methods.

\subsubsection{Implicit Regularization: Data Augmentation}
This line of work is highly relevant to regularization methods, instead of directly adding penalty terms to optimization objectives. Those studies try to augment data or add noise to training samples to mitigate model miscalibration. Label smoothing~\cite{muller2019does} and mixup~\cite{thulasidasan2019mixup} are popular approaches in this line.
 
\textit{Label smoothing}~\cite{muller2019does} soften hard labels with an introduced smoothing parameter $\alpha$ in the standard loss function (e.g., cross-entropy):
\begin{equation}
\mathcal{L}_c = - \sum_{k=1}^K y_k^s\log p_i, ~~y_k^s = y_k (1-\alpha) + \alpha/K
\end{equation}
where $y_k$ is the soft label for $k$-th category. It is shown that LS encourages the differences between the logits of the correct class and the logits of the incorrect class to be a constant depending on $\alpha$. The confidence penalty can be recovered by assuming the prior label distribution is a uniform distribution $u$ and reversing the direction of the KL divergence.
\begin{equation}
\mathcal{L}(\theta)= -\sum\log p(y|x) - \textsc{KL}(u \parallel p(y|x)).
\end{equation}

\textit{Mixup} training~\cite{thulasidasan2019mixup} is another work in this line of exploration. It studies the effectiveness of mixup~\cite{zhang2018mixup} with respect to model calibration~\cite{zhang2022and}. Mixup generates synthetic samples during training by convexly combining random pairs of inputs and labels as well. To mix up two random samples $(x_i, y_i)$ and $(x_j, y_j)$, the following rules are used:
\begin{align}
\bar{x} &= \alpha x_i + (1-\alpha) x_j \\
\bar{y} &= \alpha y_i + (1-\alpha) y_j
\end{align}
where $(\bar{x}_i, \bar{y}_i)$ is the virtual feature-target of original pairs. The authors observed that mixup-trained models are better calibrated and less prone to overconfidence in prediction on out-of-distribution and noise data. It is pointed out that mixing features alone does not bring calibration benefits; label smoothing can significantly improve calibration when used together with mixing features.

\subsubsection{Practical Considerations}
Regularization-based calibration methods differ from post-hoc approaches by directly integrating calibration objectives into the training process, enabling models to produce well-calibrated predictions without separate correction steps~\cite{kumar2018trainable,bohdal2021meta}. The computational complexity varies considerably: simple implicit regularizers like focal loss~\cite{lin2017focal} introduce negligible overhead, making them practical for large-scale training. In contrast, kernel-based methods such as Maximum Mean Calibration Error (MMCE)~\cite{kumar2018trainable} and meta-learning approaches like differentiable ECE (DECE)~\cite{bohdal2021meta} require additional computations related to kernel evaluations or meta-network optimization, which can increase training time.

Differentiable calibration proxies, including AvUC~\cite{krishnan2020improving} and MDCA~\cite{hebbalaguppe2022stitch}, offer flexible and efficient alternatives to traditional binning-based metrics by enabling end-to-end optimization of calibration errors. These methods have shown promising empirical improvements but can depend heavily on careful design and tuning to avoid overfitting~\cite{nixon2019measuring}. Overall, regularization methods provide a principled framework for calibration-aware training but may trade off simplicity and efficiency for potentially better calibration quality.

Implicit regularization via data augmentation improves calibration by enhancing generalization and reducing overfitting without modifying model architecture. Techniques like label smoothing~\cite{muller2019does} and mixup~\cite{thulasidasan2019mixup} are lightweight and easy to implement, and when combined, they often yield stronger calibration performance~\cite{zhang2022and}. However, they introduce additional hyperparameter (e.g., smoothing and mixing coefficients) that require careful tuning. These methods may also increase training time due to synthetic sample generation and can occasionally lead to underfitting if over-applied. Despite these trade-offs, they offer a practical, low-cost means to improve model calibration alongside robustness.

\subsection{Uncertainty Estimation}
This line of work aims to alleviate model miscalibration by injecting randomness.The popular methods are (1) Bayesian neural networks~\cite{blundell2015weight,fortunato2017bayesian}, (2) ensembles~\cite{lakshminarayanan2017simple}, (3) Monte Carlo (MC) dropout~\cite{gal2016dropout} and (4) Gumbel-softmax~\cite{jang2017categorical} based approaches~\cite{wang2020uncertainty,pei2022transformer}. The former three sub-categories have been discussed in recent surveys~\cite{mena2021survey,gawlikowski2021survey}. 

\subsubsection{Bayesian Neural Networks}
Given a learning objective is to minimize negative log likelihood, $\mathcal{L} = -\frac{1}{N}\sum_i^N \log p(y_i|x_i, w)$. 
The probability distribution is obtained by Softmax function as:
\begin{align}
p(y_i=m|x_i, w)=\frac{\exp(f_m(x_i, w))}{\sum_{k \in M}\exp(f_k(x_i, w)}. 
\end{align}
In the inference phase, given a test sample $x^*$, the predictive probability $y^*$ is computed by:
 \begin{align}
p(y^*|x^*, D)= \int p(y^*|x^*, w) p(w|D)dw
\end{align}
As posterior $p(w|D)$ is intractable, we perform approximation by minimizing the Kullback-Lieber (KL) distance. 
This can also be treated as the maximization of \textit{ELBO}:
\begin{align}
\mathcal{L_\theta}=\int q_\theta(w)p(Y|X, w)dw-\textsc{KL}[q_\theta(w)\parallel p(w)]
\label{eq:bayesian_inference}
\end{align}
where $\theta$ are the variational parameters. With the re-parametrization trick~\cite{kingma2015variational}, a differentiable mini-batched Monte Carlo (MC) estimator can be obtained. 

The uncertainty estimation can be done by performing $T$ inference runs and averaging predictions:
$
p(y*|x*)= \frac{1}{T}\sum_{t=1}^T p_{w_t}(y^*|x^*, w_t).
$

\subsubsection{Stochastic and Gradient-Based Uncertainty Methods}
Several uncertainty estimation techniques improve calibration by introducing stochasticity or reweighting gradients during training. MC Dropout~\cite{gal2016dropout}, deep ensembles~\cite{lakshminarayanan2017simple,zou2023adaptive}, and Gumbel-softmax sampling~\cite{jang2017categorical,wang2020uncertainty} inject randomness in different ways: via dropout masks, independently trained model weights, or sampled attention distributions, respectively. These methods typically rely on multiple forward passes ($T$ samples) to approximate predictive uncertainty.

More recently, Lin et al.~\citeyear{lin2025uncertainty} proposed a sample-weighted gradient approach, BSCE-GRA, which scales gradients based on Brier Scores across all logits. Unlike traditional loss weighting, this method directly adjusts gradient magnitudes to prioritize uncertain samples without hindering convergence. Empirical results show that BSCE-GRA achieves competitive calibration while preserving model accuracy.

\subsubsection{Practical Considerations}
Uncertainty estimation methods enhance calibration by modeling predictive confidence through stochastic techniques like Bayesian neural networks~\cite{blundell2015weight}, MC Dropout~\cite{gal2016dropout}, ensembles~\cite{lakshminarayanan2017simple}, and Gumbel-softmax sampling~\cite{jang2017categorical}. These approaches capture both aleatoric and epistemic uncertainty and improve reliability in high-risk settings. Gradient-based methods such as BSCE-GRA~\cite{lin2025uncertainty} further refine training by prioritizing uncertain samples without modifying model architecture. However, these methods often require multiple forward passes or model instances, increasing computational and memory costs. This limits their practicality in real-time or resource-constrained environments. Despite these trade-offs, they provide robust calibration and uncertainty quantification when efficiency is not the primary constraint.

\subsection{Hybrid Calibration Methods}
\label{sec:hybird_calibration}
Beside applying each  method independently, we can always have calibration compositions by combining two or more methods. One straightforward way to combine non-post-hoc methods with post-hoc methods. For instance, performing Temperature Scaling (TS) after employing the regularization method and implicit calibration~\cite{kumar2018trainable,bohdal2021meta}. Thulasidasan
et al.~\citeyear{thulasidasan2019mixup} observed that the combination of label smoothing and mixup training significantly improved calibration. While there are several possibilities for combining different approaches, we highlight some interesting compositions.
 
\subsubsection{Ensemble Temperature Scaling (ETS)}
Zhang et al.,~\citeyear{zhang2020mix} gave three important definitions related to calibration properties: accuracy-preserving, data-efficient, and expressive. They pointed out that TS is an accuracy-preserving and data-efficient approach but is less expressive. Ensemble Temperature Scaling (ETS) was proposed to improve TS expressivity while maintaining the other two properties:
\begin{equation}
T(z;w,\tau)=w_1 T(z;\tau)+w_2z+w_3K^{-1}
\end{equation}
There are three ensemble components: the original TS $T(z;\tau)=(z_1^{\tau^{-1}},...,z_k^{\tau^{-1}})/\sum_{k=1}^Kz_k^{\tau^{-1}}$, uncalibrated prediction with $\tau=1$ and uniform prediction for each class $z_k=K^{-1}$.

\subsubsection{Temperature Scaling with MC Dropout}
Laves et al.~\citeyear{laves2019well} extended TS to dropout variational inference to calibrate model uncertainty. The key idea is to insert $\tau$ before final softmax activation and insert TS with $\tau>0$ before softmax activation in MC integration:$
\hat{p} = \frac{1}{N}\sum_{i=1}^N\textsc{softmax}(\frac{f_{w_i}(x)}{\tau})$
where $N$ forward passes are performed to optimize $\tau$ with respect to NLL on the validation set. Then the entropy of the softmax likelihood is used to represent the uncertainty of all $C$ classes.
\begin{equation}
H(p) = - \frac{1}{\log C}\sum p^{c}\log p^c, H \in [0, 1]
\end{equation}

\subsubsection{Conformal Prediction with MC Dropout}
Cocheteux et al.~\citeyear{cocheteux2025uncertainty} propose an uncertainty-aware online extrinsic calibration pipeline that integrates MC Dropout for sampling epistemic uncertainty over calibration parameters with Conformal Prediction~\cite{angelopoulos2021gentle,vovk2020conformal,vovk2020computationally} to generate prediction intervals with statistically guaranteed coverage. They apply MC‑Dropout inference on backbone calibration models, compute nonconformity scores from the dropout-generated samples, and use conformal quantiles to construct confidence intervals. Evaluations on multiple datasets demonstrate that the method achieves near-nominal coverage (PICP), acceptable interval widths (MPIW), and improved interval scores (IS), all while supporting dynamic, online calibration scenarios.

\subsubsection{Practical Considerations}
Combining different types of calibration methods—particularly post-hoc techniques with regularization-based or uncertainty-aware methods—can yield more robust and flexible calibration pipelines. Such hybrid strategies often leverage complementary strengths: for instance, the simplicity and data efficiency of post-hoc methods like Temperature Scaling (TS) alongside the expressiveness or generalization benefits of training-time techniques. However, these benefits may come at the cost of increased model complexity, interpretability challenges, and higher computational overhead. For example, Ensemble Temperature Scaling (ETS)~\cite{zhang2020mix} enhances expressivity but introduces additional parameters and components that must be tuned. Similarly, integrating Monte Carlo (MC) Dropout into post-hoc calibration (as in TS or Conformal Prediction settings) improves uncertainty estimation but requires multiple stochastic forward passes, potentially increasing inference time. Thus, while hybrid methods present promising improvements in calibration quality, their design should carefully consider trade-offs between performance, scalability, and deployment constraints.

\subsection{Discussion and Analysis}
We have introduced some representative methods for each line of methodology categorization. In this section, we analyze and describe their strengths and limitations. Additionally, we outline the popular benchmark datasets and backbone models widely used in the literature.

\subsubsection{Summary of Strengths and Limitations}
Table~\ref{tab:calibration_strengthen_limitation} presents the strengths and limitations of each calibration category. In general, different calibration categories have their own advantages. For methods beyond Post-hoc calibration, the limitations mostly revolve around increased model architectural and computational complexity. While an exact comparison in this regard depends on the specific method selected, a rough complexity ranking could be hybrid method, uncertainty estimation, regularization, and post-hoc method, in descending order of complexity.
\begin{table}[htb]
  \centering
  \footnotesize
  \begin{tabular}{p{3cm}p{5.75cm}p{5.75cm}}
    \hline
    Calibration Methods & Strengthens & Limitations \\ \hline
    Post-hoc Calibration &
      (1) Simple and effective: this line of work can be easily implemented and integrated into existing models. On the other hand, it shows its effectiveness and accuracy-preserving~\cite{zhang2020mix} with highly competitive predictive and calibration performance. &
      (1) Suboptimal performance: Post-hoc calibration techniques may not fully optimize the calibration of the model, potentially leading to suboptimal calibration performance. Additionally, Chidambaram et al.,~\citeyear{chidambaram2024on} recently suggested that the performance of temperature scaling degrades with the amount of overlap between classes.
\\ \hline
    Regularization &
      (1) Training-time calibration, as calibration is performed during the training, we can obtain a well calibrated model directly after training, omitting the post-hoc steps. \newline
      
      (2) Alleviating overfitting: regularization methods improves calibration through alleviated overfitting such as L2 regularization and Focal loss~\cite{lin2017focal}. Thus this line of methods can also improve generalization performance. 
 &
      Implementation and Tuning Complexity. Implementing regularization method is less straightforward as compared to post-hoc ones. Also this line of work usually introduce hyper-parameters, for example, smoothing factor in label smoothing~\cite{muller2019does} and ~\cite{mukhoti2020calibrating} empirically showed that the importance of $\gamma$ in calibrating deep models. However, selecting the appropriate hyper-parameters can be challenging and may require extensive hyper-parameter tuning.  \\
    \hline
    Uncertainty Estimation (UE) &
    
    (1) Uncertainty quantification. Bayesian-method based calibration can quantify the uncertainty in the calibrated model~\cite{muehleisen2016bayesian}.  
    
    (2) Incorporating prior information. UE methods allow incorporation of prior knowledge or beliefs about the parameters being estimated and provide more robust estimates, especially when dealing with limited data~\cite{kennedy2001bayesian}.
 &
      Computational complexity. Many uncertainty estimation methods involve computationally intensive procedures, which can be impractical for large datasets or real-time applications, limiting their scalability. For example MC-dropout~\cite{gal2016dropout} requires multiple inference runs and perform averaging on predictions. \\ \hline
    Hybrid Methods &

      Combines strengths: Hybrid methods can leverage the advantages of multiple approaches, potentially leading to better overall performance by addressing the limitations of individual techniques. Section~\ref{sec:hybird_calibration} describes some of these methods &

      Increased complexity: Hybrid methods usually introduce additional complexity in model calibration, Integrating multiple techniques into a cohesive framework may pose implementation challenges and increase the risk of errors requiring more extensive validation and testing to ensure robustness and effectiveness. \\
    \hline
  \end{tabular}
    \caption{The summarization of strengthens and limitations of each line of calibration methods.}
  \label{tab:calibration_strengthen_limitation}
\end{table}

\subsubsection{Backbone Models and Datasets}
In the literature, the datasets and backbone models (the models to be calibrated) used in calibration methods vary from method to method, posing a challenge for future methods in making comparisons with previous ones. Here, we outline some representative datasets and backbone models that have been utilized in previous works to facilitate future baseline selections.
\begin{itemize}
    \item \textbf{Backbone Models}. The widely used backbone models include: LeNet-5~\cite{lecun1998gradient}, ResNet~\cite{he2016deep}, Fully-connected convolution network (FCN)~\cite{long2015fully}, WideResNet~\cite{zagoruyko2016wide}, DenseNet~\cite{huang2017densely}, ResNext~\cite{xie2017aggregated}, Global-pooling CNN (GP-CNN)~\cite{lin2013network}, LSTM~\cite{hochreiter1997long}. 
    \item \textbf{Datasets}. The widely used vision datasets are CIFAR10/100~\cite{krizhevsky2009learning}, Cambridge-driving Labeled Video Database (CamVid)~\cite{brostow2008segmentation}, CoCo~\cite{lin2014microsoft}, LPBA40~\cite{shattuck2008construction}, Caltech-UCSD Birds (CUB)~\cite{welinder2010caltech,wah2011caltech},Stanford Cars~\cite{krause20133d}, SVHN~\cite{netzer2011reading}, Tiny-ImageNet~\cite{Le2015TinyIV}, STL-10~\cite{coates2011analysis}, Pascal-Voc-2012~\cite{everingham2010pascal}, MINST and Fashion-MNIST~\cite{xiao2017fashion}; NLP datasets such as MR~\cite{pang2005seeing}, TREC~\cite{li2002learning}, Stanford Sentiment Treebank (SST)~\cite{socher2013recursive}, 20Newsgroups~\cite{lang1995newsweeder}, IMDB~\cite{maas-EtAl:2011:ACL-HLT2011}, Penn Treebank~\cite{marcus1993building}, WikiText-2~\cite{merity2016pointer}
\end{itemize}

Table~\ref{tab:methods_datasets} outlines the categorization, datasets and backbone models to be calibrated. As we can see, for backbones, ResNet~\cite{he2016deep}, WideResNet~\cite{zagoruyko2016wide} and DenseNet~\cite{huang2017densely} are most popular ones for image classification task. And LSTM~\cite{hochreiter1997long} and GP-CNN~\cite{lin2013network} are widely used in NLP tasks. Regarding datasets, mostly all selected methods used CIFAR10/100~\cite{krizhevsky2009learning} for image classification, and 20Newsgroups~\cite{lang1995newsweeder}, IMDB~\cite{maas-EtAl:2011:ACL-HLT2011}
\begin{table}[htb]
\setlength{\tabcolsep}{3pt}
\scriptsize
\begin{tabular}{p{4cm}lp{2.5cm}p{3.5cm}p{3cm}} 
\hline
Methods & Categorization & Measurement & Backbone Models & Datasets \\ \hline
Dirichlet Calibration~\cite{kull2019beyond} & Post-hoc & ECE,NLL,MCE, CECE, Bier &  ResNet-110/110SD/152SD, DenseNet-40, WisdeNet-32, LeNet-5&CIFAR10/100, SVHN \\
ATS~\cite{mozafari2018attended} & Post-hoc & ECE,NLL& ResNet52, WideResNet-32, DenseNet-40/100, VGG-16 & CIFAR10/100, MINST \\
BTS~\cite{ji2019bin} & Post-hoc & ECE & CNN,ResNet-50/110, WideResNet-28/10, DenseNet-100, VGG-16 & CIFAR10/100, CUB, Stanford Cars  \\
LTS~\cite{ding2021local} & Post-hoc & ECE,MCE,ACE, SCE&  FCN, ResNet-101 & CoCo, CamVid, LPBA40\\
FLSD~ ~\cite{mukhoti2020calibrating} & Regularization &ECE,MCE,NLL, ACE, CECE& ResNet-50/110, Wide-ResNet-26-10, DenseNet-121 &CIFAR10/100, SST, 20Newsgroups\\
MMCE~\cite{kumar2018trainable} & Regularization & ECE, NLL, Brier& ResNet-50/110, Wide-ResNet-28-10, Inception-v3, Global Pooling CNN, LSTM, Tree-LSTM & CIFAR10/100, 20Newsgroups,IMDB, SST, Birds CUB 200  \\
Meta-Calibration~\cite{bohdal2023meta} & Regularization & ECE,NLL,ACE, CECE& ResNet-18/50/110, WideResNet-26-10, GP-CNN & CIFAR10/100,SVHN, 20Newsgroups\\
Margin-based LS~\cite{liu2022devil} & Regularization & ECE, ACE & ResNet-50/101,GP-CNN &CIFAR10,Tiny-ImageNet,Pascal-Voc-2012,CUB, 20Newsgroups\\
Mix-Up~\cite{thulasidasan2019mixup} & Regularization & ECE, OE & ResNet-18/50, ResNext-101, VGG-16& CIFAR10/100,STL-10,ImageNet,Fashion-MNIST,MR, TREC, IMDB\\
Inverse Focal Loss~\cite{wang2021rethinking} & Regularization  & ECE & ResNet-32, GP-CNN&CIFAR10/100, SVHN, 20Newsgroups\\

SWA-Gaussian~\cite{maddox2019simple} & Uncertainty  & NLL & WideResNet-28-10, DenseNet-161, Resnet-152,PreResNet-164, VGG-16,  LSTM &CIFAR10/100, STL-10, ImageNet, Penn Treebank, WikiText-2\\
Ensemble~\cite{rahaman2021uncertainty} & Uncertainty  & ECE, NLL, Bier &  ResNet-18/34, LeNet-5 &CIFAR10/100, MNIST\\
Mix-n-Match~\cite{zhang2020mix} & Hybird & ECE & ResNet-110,DenseNet-40,WideResNet-28-10, WideResNet-52-2, LeNet-5, VGG-19, ResNext-101 &CIFAR10/100, ImageNet\\

TS+MC dropout~\cite{laves2019well} & Hybird & ECE, UCE & ResNet-18/101, DenseNet-169 &CIFAR10/100\\ \hline
\end{tabular} 
\caption{The calibration methods and their categorization, datasets and backbone models to be calibrated with those methods. }
\label{tab:methods_datasets}
\end{table}

\section{Calibrating Pre-trained Large Models}
Pre-trained large models, including vision, language, or vision-language models, have been increasingly used in many safety-critical and customer-facing applications. The calibration of those large models has been recently studied and revisited~\cite{Minderer2021,levine2023enabling}.
 
\subsection{Large Vision Models (LVMs)}
Minderer et al.~\citeyear{Minderer2021} studied recent state-of-the-art vision models that include vision transformer~\cite{dosovitskiyimage} and MLP-mixer~\cite{tolstikhin2021mlp}. They found out that the model size and amount of pre-training in the recent model generation could not fully explain the observed decay of calibration with distribution shift or model size in the prior model generation. They also discussed the correlation between in-distribution and out-of-distribution (OOD) calibration. They pointed out that the models with better in-distribution calibration also gave better calibration on OOD benchmarks. LeVine et al.~\citeyear{levine2023enabling} studied the calibration of CLIP~\cite{radford2021learning} as a zero-shot inference model and found that CLIP is miscalibrated in zero-shot settings.They showed effectiveness of learning a temperature on an auxiliary task and applying it to inference regardless of prompt or datasets.

\subsection{Large Language Models (LLMs)}
Pre-trained LLMs like GPT-3 and its successors exhibit strong generalization across diverse tasks through in-context learning (ICL). However, their confidence estimates are often poorly aligned with actual correctness, especially in few-shot and zero-shot settings. Unlike traditional supervised models, LLMs are typically used in prompt-based or API-only scenarios where internal logits are not directly accessible, making black-box calibration a key research challenge. This has motivated a growing body of work aimed at reducing overconfidence, mitigating bias, and improving the reliability of LLM predictions across task types and domains.

The outcome of prompting-based learning can be unstable and introduce biases through prompt format and example selection~\cite{zhao2021calibrate}. The introduced biases include majority label bias, recency bias, and token bias. To mitigate this, \cite{zhao2021calibrate} proposed \textit{contextual calibration}, which improves the predictive power of GPT-3 in few-shot settings by estimating bias using a context-free input such as "N/A", and then applying vector scaling~\cite{guo2017calibration}:
\begin{align}
\hat{p} = p(y[``N/A"], \textsc{Context}),~~p_{calibrated} = W\hat{p}
\end{align}
where $W$ is a diagonal matrix.

Domain-Context Calibration (DC) ~\cite{fei2023mitigating} was proposed to mitigate label biases in ICL. Instead of a fixed token like "N/A", the method uses random in-domain text to estimate label-wise priors:
\begin{align}
\hat{p} = \frac{1}{T}\sum_{t=1}^T p(y[\textsc{In-Domain Random text}]_t, \textsc{Context})\\
p_{calibrated} = p(y|x, \textsc{Context}) - \hat{p}
\end{align}

Zheng et al.~\citeyear{zheng2023large} investigated selection bias in multiple-choice tasks and identified token bias as a core factor. Their proposed method, \textit{PriDe}, is a label-free, inference-time debiasing method with strong efficiency and cross-domain generalization.

Park et al.~\citeyear{park2022calibration} extended mixup~\cite{zhang2018mixup} training for pre-trained LMs by generating synthetic examples based on the Area Under the Margin (AUM), improving calibration in classification settings. Another approach called \textit{Prototypical calibration (PROCA)}~\cite{han2022prototypical} emphasizes the importance of decision boundaries. It fits a Gaussian mixture model (GMM) over model outputs:
\begin{align}
P_{GMM}(\hat{p}) = \sum_{k=1}^K \alpha_k p_G(\hat{p}|u_k,\Sigma_k), ~~~
p_{calibrated} = \frac{1}{T} \sum_{t=1}^T P_{GMM}(\hat{p}_t)
\end{align}
where parameters $\{\alpha_k, u_k, \Sigma_k\}_{k=1}^K$ are estimated via Expectation-Maximization~\cite{moon1996expectation}.

Zhou et al.~\citeyear{zhou2023batch} identified limitations in earlier methods. For example, contextual and domain-context calibration struggle with multi-sentence inputs, and PROCA may overfit. They proposed \textit{Batch Calibration}, which uses class-wise prior estimation over batches to correct contextual bias. Unlike prior techniques that rely on handcrafted calibration prompts or context-free inputs (e.g., "N/A"), Batch Calibration estimates contextual bias directly from a batch of real task inputs. Specifically, it marginalizes model outputs across a batch to estimate the background distribution over labels, then debiases each individual prediction accordingly:
\begin{align}
\hat{p}(y) = \frac{1}{B} \sum_{i=1}^B p(y | x_i, \textsc{Context}), \quad
p_{\text{calibrated}}(y|x) = p(y | x, \textsc{Context}) - \hat{p}(y)
\end{align}
Here, $B$ denotes the batch size. The method is robust to prompt variations. It unifies previous approaches such as Contextual Calibration and Domain Calibration under a more stable framework, and achieves state-of-the-art calibration and accuracy across multiple tasks using models like PaLM-2 and CLIP.

Shen \textit{et al.}~\citeyear{shen2024thermometer} introduced \textit{Thermometer}, an auxiliary‑model‑based, post‑hoc calibration method designed specifically for open‑form large language models (LLMs). Unlike common approaches that rely on multiple forward passes or fine‑tuning at inference, Thermometer trains a lightweight, task-agnostic calibration network across multiple datasets. At inference time, this auxiliary model adjusts the LLM’s predicted probabilities to improve ECE without altering the LLM’s outputs or requiring additional expensive inferences. Empirical results on multiple benchmarks show that Thermometer consistently achieves lower calibration error than temperature scaling or MC‑based methods, preserves accuracy, and provides faster inference—thanks to its single-pass inference overhead.

Zhang et al.~\citeyear{zhang2024atomic} proposed \textit{atomic calibration}, a novel framework that decomposes long-form answers into individual, self-contained factual units known as atomic claims, and then estimates a separate confidence score for each. Two types of methods are introduced: generative methods that assess confidence based on sampling consistency across multiple generations, and discriminative methods that query the model directly for confidence in each claim. The authors also introduce new evaluation metrics such as Unnormalized and Quantized Claim Calibration Errors (UCCE and QCCE) to measure calibration quality at both the claim and response levels. Experiments conducted across several open-source LLMs and long-form QA datasets show that atomic calibration yields significantly more accurate and interpretable confidence estimates. Furthermore, combining generative and discriminative signals produces the best overall calibration performance. This work demonstrates the necessity of fine-grained calibration techniques for open-form LLM outputs and provides a scalable framework to achieve it.


\section{Calibration Applications}
In this section, we describe the application of calibration to different relevant domains. 
\subsection{Medical Imaging and Analysis}

Calibrated deep learning models play a crucial role in medical imaging tasks such as disease diagnosis and medical image analysis~\cite{ricci2023towards}. These models provide accurate predictions and uncertainty estimates, assisting clinicians in making informed decisions about patient care. For instance, calibrated models improve the reliability of automated diagnoses in radiology, pathology, and other medical imaging domains. By aligning predicted probabilities with observed outcomes, these models enhance the trustworthiness of automated medical decision support systems. Numerous studies have demonstrated the effectiveness of model calibration techniques in improving the accuracy and interpretability of medical imaging models. Penso et al. ~\citeyear{10399826} presents a network calibration procedure to make be robust to label noise based on observation that the confusion matrix of the noisy labels can be expressed as the matrix product between the confusion matrix of the clean labels and the label noises. Ricci Lara
et al.~\citeyear{ricci2023towards} identified the issue of discrimination and calibration biases in models trained for
automatic detection of malignant dermatological conditions from skin lesions images. They found that calibration metrics may appear to show that model outputs do not fairly represent uncertainty.

\subsection{Financial Decision Making}
In the finance industry, calibration can provide more reliable access to financial risk assessment such as fraud detection~\cite{bahnsen2014improving,habibpour2023uncertainty}, and financial forecasting~\cite{islyaev2015electricity}. These models produce accurate predictions of asset prices, market trends, and financial risks, assisting investors and financial institutions in decision-making and risk management. For example, calibrated models help financial institutions identify fraudulent transactions more effectively by providing reliable estimates of transaction risk. Additionally, calibrated models improve the accuracy of financial forecasting models, enabling investors to make informed decisions about portfolio allocation and asset management. Several studies have demonstrated the effectiveness of model calibration techniques in improving the reliability and accuracy of financial forecasting models.

\subsection{Autonomous Driving}
In the domain of autonomous vehicles, calibrating model predictions is essential for safe and reliable navigation in complex traffic scenarios. These models enable accurate perception, prediction, and decision-making, ensuring that self-driving vehicles can operate safely in dynamic environments. For example, ~\cite{peng2021uncertainty} applied MC-dropout~\cite{gal2016dropout} to quantify uncertainty of object detection and performed calibration based on the uncertainty analysis. 

\section{Conclusion and Future Work}
\label{sec:future}

We have reviewed the state-of-the-art calibration methods, described with the motivations, causes, measurement metrics, and categorizations. Then we discussed the details and principles of recent methods as well as their individual advantages and disadvantages. Despite recent advances in calibrating deep models, there are still some challenges and under-explored aspects and needs further exploration.

\subsection{Mitigating Calibration Bias}
\begin{figure}[htb]
	\centering
  	\subfloat{{\includegraphics[width=0.25\textwidth]{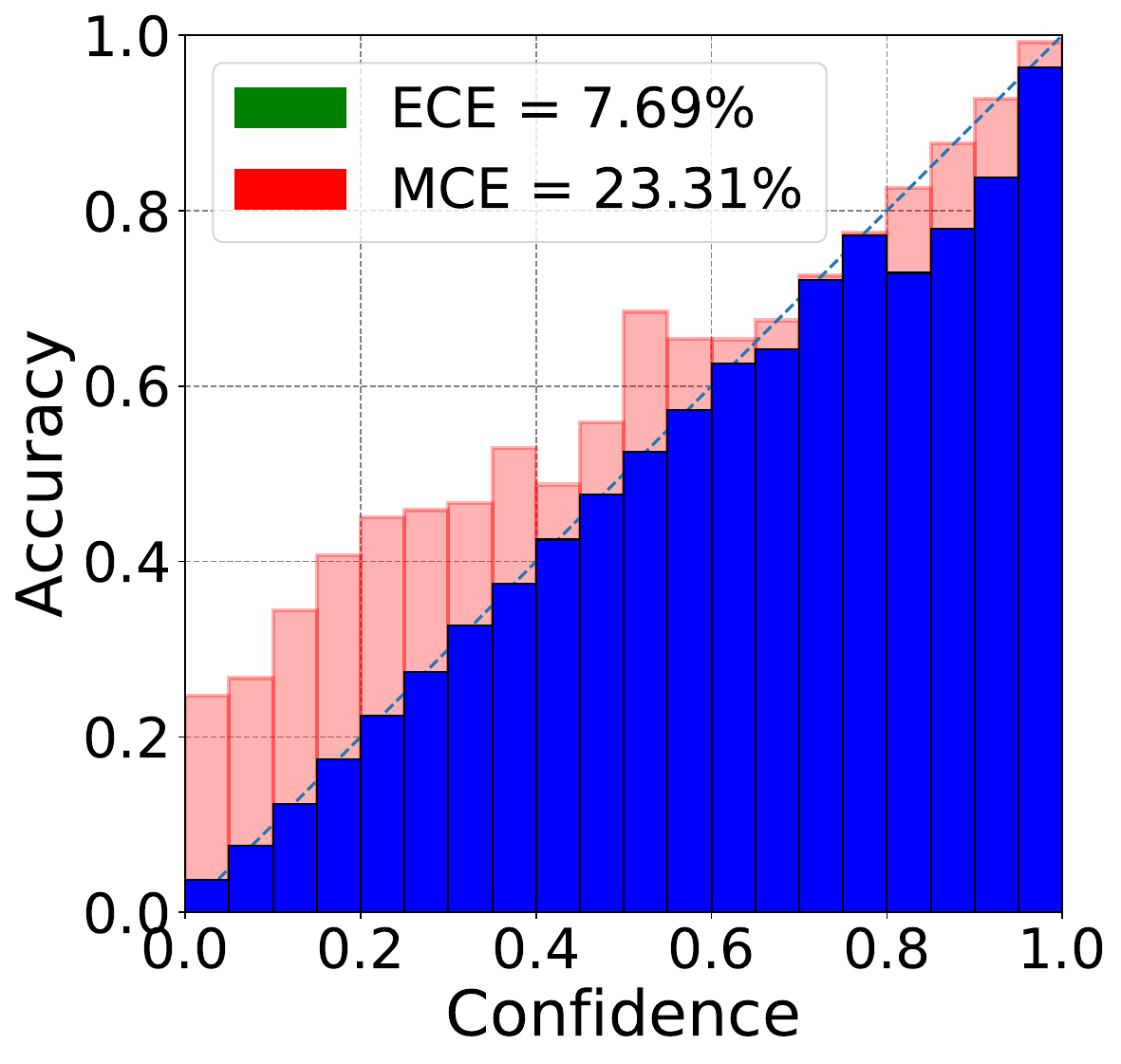} }}
  	 	\subfloat{{\includegraphics[width=0.25\textwidth]{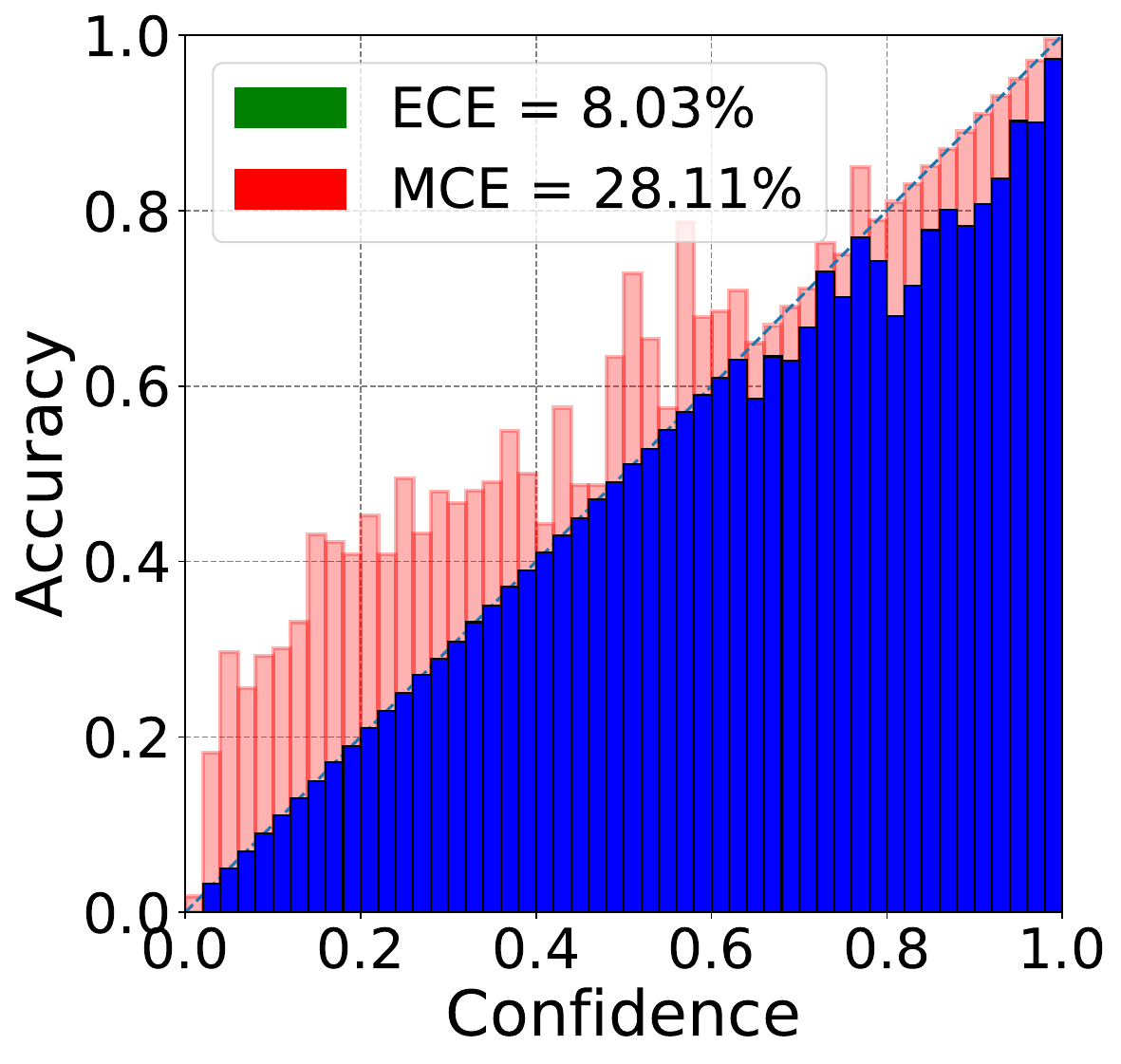} }}
  	 	\subfloat{{\includegraphics[width=0.25\textwidth]{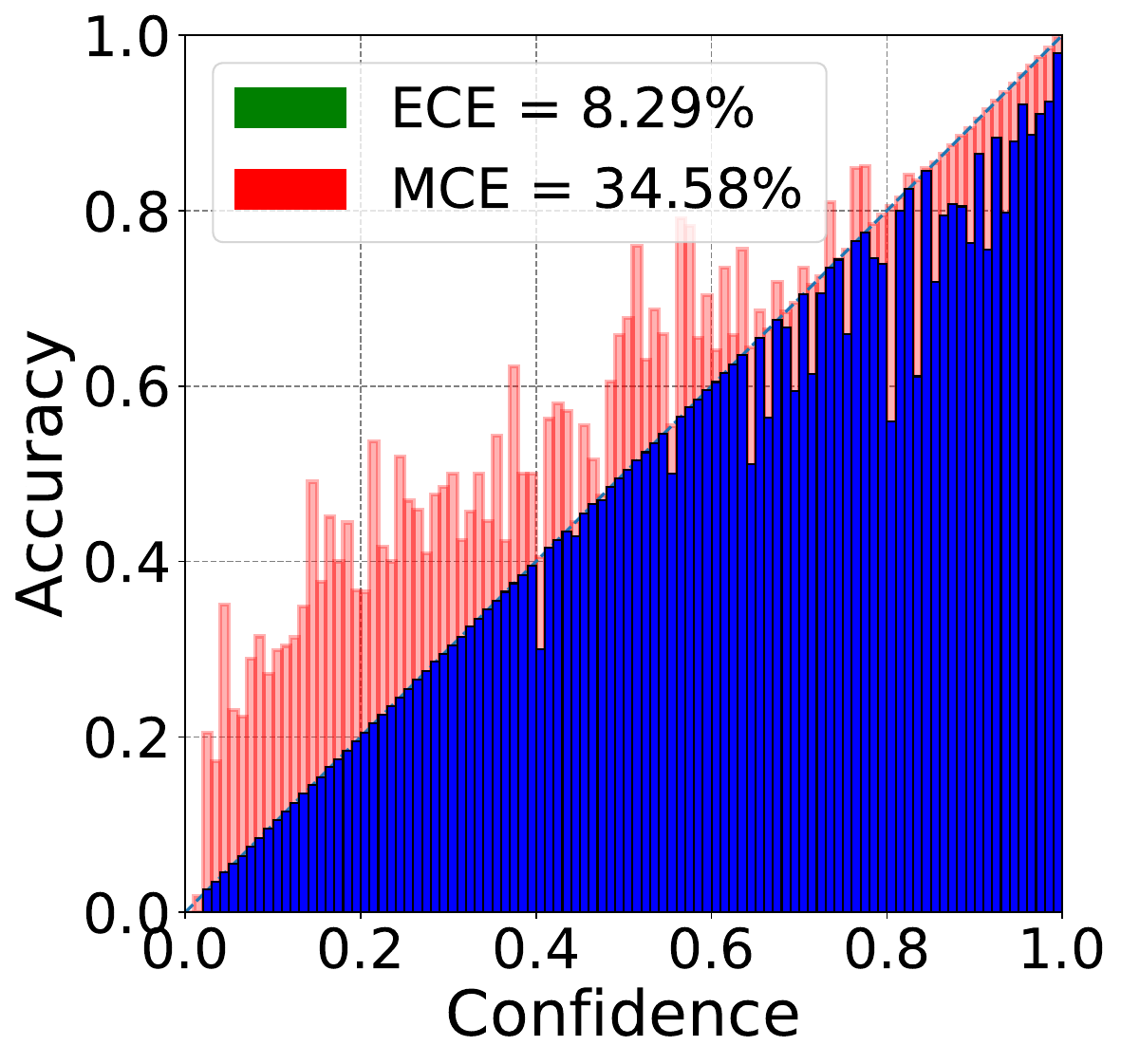} }}
	\caption{The reliability diagrams for a model trained on CIFAR100 with different bin numbers (left to right: 20, 50, 100). The diagonal dash presents perfect calibration, the red bar presents the gap to perfect calibration on each bin.The calibration error is sensitive to increasing bin numbers.}
\label{fig:rc_c100_calibration_more} 
\end{figure}

Accurately and reliably measuring calibration is still challenging due to the introduced biases from the binning mechanism and the finite sample numbers~\cite{minderer2021revisiting}. For the former challenge, it mainly suffers from sensitivity and data inefficiency issues. The sensitivity to the binning scheme is presented in Figure~\ref{fig:rc_c100_calibration_more}. We can see that for a given model, increasing bin numbers gives higher ECE and MCE scores. A KDE-based ECE Estimator~\cite{zhang2020mix} was proposed to replace histograms with non-parametric density estimators, which are continuous and more data-efficient. Measuring the bias is then important for having a correct calibration evaluation. Roelofs et al.~\citeyear{roelofs2022mitigating} proposed Bias-by-Construction (BBC) to model the bias in bin-based ECE as a function of the number of samples and bins. It confirms the existence of non-negligible statistical biases. To follow this line of work, future efforts will include developing an unbiased calibration estimator, exploring the trade-off between calibration bias and variance as mentioned in ~\cite{roelofs2022mitigating}.

\subsection{Calibrating Generative Models}
Recent efforts in model calibration have predominantly centered around classification and regression tasks, leaving a noticeable gap in discussions regarding calibration for sequence generation. While advancements in calibration techniques have greatly benefited these tasks, the applicability and efficacy of such techniques in sequence generation remain under-explored. Building upon the seminal work of ~\cite{kumar2019calibration}, which highlighted the inadequately calibrated token-level probabilities in neural machine translation tasks, there is a pressing need to extend these insights to a broader range of generative models. The findings of~\cite{kumar2019calibration,zhao2022calibrating} revealed a crucial connection between token-level probability miscalibration and the counter-intuitive drop in BLEU scores with increased beam size, as noted in~\cite{koehn2017six}. Furthermore, they demonstrated that improved sequence-level calibration can be achieved through the re-calibration of token-level probabilities. This underscores the importance of addressing token-level probability miscalibration not only for enhancing performance metrics but also for improving the overall reliability and coherence of generated sequences.

Building on these insights, recent studies such as Zhao et al.~\citeyear{zhao2021calibrate} have demonstrated that \textit{token-level probability miscalibration} is a persistent issue in Large Language Models (LLMs), particularly in few-shot learning scenarios. One key contributor is the \textit{softmax bottleneck}—as identified in \cite{yang2018breaking}—which becomes especially problematic in models with large vocabularies. In sequence generation tasks, early-stage miscalibrations can compound across tokens, leading to amplified errors throughout the generated output.

Looking ahead, developing methods to \textit{accurately calibrate token-level probabilities} in generative settings is an important and emerging direction, particularly in the context of LLMs. Unlike classification or regression, sequence generation introduces unique challenges that demand novel calibration techniques. Recent efforts such as \textit{generative calibration}~\cite{jiang2023generative}, \textit{sequence likelihood calibration}~\cite{zhao2022calibrating}, and its enhanced variant \textit{SLIC}~\cite{zhao2023slic} offer promising starting points for advancing calibration in this space.

\newpage
\newpage

\vskip 0.2in
\bibliography{sample}
\bibliographystyle{theapa}

\end{document}